\documentclass[10pt,twocolumn,letterpaper]{article}

\usepackage[pagenumbers]{cvpr} 

\usepackage{graphicx}
\usepackage{amsmath}
\usepackage{amssymb}
\usepackage{booktabs}

\usepackage{algorithmic}
\usepackage{algorithm}
\usepackage[accsupp]{axessibility}

\usepackage[pagebackref,breaklinks,colorlinks]{hyperref}
\usepackage{pifont}

\usepackage[capitalize]{cleveref}
\crefname{section}{Sec.}{Secs.}
\Crefname{section}{Section}{Sections}
\Crefname{table}{Table}{Tables}
\crefname{table}{Tab.}{Tabs.}

\def\cvprPaperID{13} 
\def\confName{CVPRW}
\def\confYear{2022}

\newcommand{\cN}{\mathcal{N}}

\newcommand{\cT}{\mathcal{T}}

\newcommand{\cC}{\mathcal{C}}

\newcommand{\cV}{\mathcal{V}}
\newcommand{\cE}{\mathcal{E}}

\newcommand{\cG}{\mathcal{G}}
\newcommand{\xmark}{\ding{55}}%
\newcommand{\cmark}{\ding{51}}%

\begin{document}

\title{Multi-Camera Multiple 3D Object Tracking on the Move \\ for Autonomous Vehicles}

\author{Pha Nguyen$^{1}$, Kha Gia Quach$^{2}$, Chi Nhan Duong$^{2}$, Ngan Le$^{1}$, Xuan-Bac Nguyen$^{1}$, Khoa Luu$^{1}$ \\
    $^{1}$ CVIU Lab, University of Arkansas, USA \quad
    $^{2}$ Concordia University, CANADA \\
    \tt\small$^{1}$\{panguyen, thile, xnguyen, khoaluu\}@uark.edu \quad
    $^{2}$\{dcnhan, kquach\}@ieee.org}

\maketitle

\begin{abstract}
    The development of autonomous vehicles provides an opportunity to have a complete set of camera sensors capturing the environment around the car. Thus, it is important for object detection and tracking to address new challenges, such as achieving consistent results across views of cameras. To address these challenges, this work presents a new Global Association Graph Model with Link Prediction approach to predict existing tracklets location and link detections with tracklets via cross-attention motion modeling and appearance re-identification. This approach aims at solving issues caused by inconsistent 3D object detection. Moreover, our model exploits to improve the detection accuracy of a standard 3D object detector in the nuScenes detection challenge. The experimental results on the nuScenes dataset demonstrate the benefits of the proposed method to produce SOTA performance on the existing vision-based tracking dataset.

\end{abstract}

\section{Introduction}

Object detection and tracking have become one of the most important tasks in autonomous vehicles (AV). Recent development of deep learning methods has dramatically boosted the performance of object understanding and tracking in autonomous driving applications thanks to the availability of public datasets. Far apart from prior video tracking datasets collected via single or stereo cameras, e.g., KITTI \cite{geiger2012we}, recent public datasets and their defined tracking problems have become more realistic with multiple cameras in autonomous vehicles. They usually have a full set of camera sensors that aim to create a 360$^\circ$ surround view and provide more redundancy as backup, i.e. more overlapping field-of-views.
There are some popular large-scale tracking datasets with multiple sensor setup, such as nuScenes \cite{caesar2020nuscenes}, Waymo \cite{sun2019scalability}, Lyft \cite{skeete2018level}, or Argoverse \cite{chang2019argoverse}. They have a lot more data than KITTI ranging from multiple surrounding cameras, LiDAR, radars and GPS.

\begin{figure}
    \centering
    \includegraphics[width=0.95\linewidth]{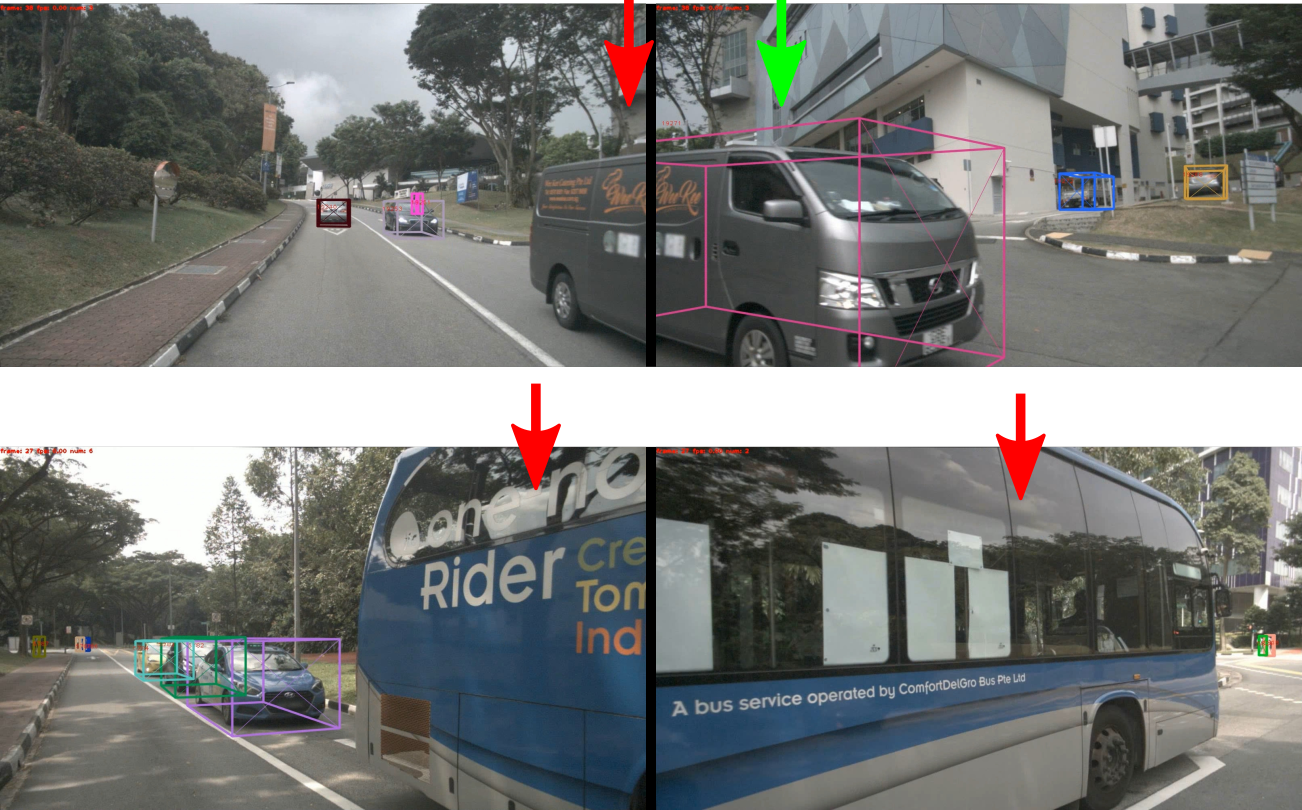}%
    \caption{First row: the object detector and tracking method DEFT \cite{Chaabane2021deft} fails to detect partial objects in one camera but can detect in another camera, Second row: The detector fails to detect objects in both cameras. Green arrow indicates true positive detection sample, red arrows indicate false negative detection samples.}
    \label{fig:failed_detection_case}
\end{figure}

Having enormous data as in recent public datasets helps to improve deep learning based 3D object detection. However, it also poses more challenging problems in practice, such as maintaining high accuracy and latency performance in variety points of views and environments.
In addition, Multiple Object Tracking (MOT) is usually employed together with 3D object detection to track objects and maintain stability of prediction across video frames. In order to handle multiple views, a common approach to Multi-Camera Multiple Object Tracking (MC-MOT) \cite{cai2014exploring, chen2016equalized} is to firstly apply an MOT approach on each camera independently, i.e. single camera tracking (SCT), then link local tracklets across cameras together via global matching steps based on Re-ID features.
However, this approach creates more errors, i.e. fragmented local tracklets, and more computation since the data association and the matching steps will perform multiple times both locally and globally. Therefore, using SCT multiple times is not the optimal option. In addition, it is unable to handle scenarios when the detector fails to detect objects from one of the cameras as shown in Fig. \ref{fig:failed_detection_case}.

Therefore, this work proposes to formulate MC-MOT problem as a \emph{global association graph} in a $360^{\circ}$ view using an object detection as the inputs instead of SCT trajectories. Our proposed MC-MOT approach not only models object motion but also the appearance of each tracked object. We encode both location and appearance features in the node embeddings of the proposed graph where the nodes corresponding to each tracked object are updated and added to the graph over time. In addition, we adopt the new self-attention and cross-attention layers to decode motion and location, then propagate them across camera systems via 3D-to-2D transformation.

\noindent
\textbf{Contributions of this Work. }
The main contributions of this work can be summarized as follows. A new MC-MOT framework is firstly introduced where a global graph is constructed with \textit{nodes} containing both appearance and motion features of the tracked objects and \textit{the weighted edges} between tracked objects or nodes. The edge weights are computed based on the similarity in appearance and location between two tracked objects or nodes. Secondly, we present a new \emph{Auto-regressive Graph Transformer network} including a self-attention layer to transform appearance features and cross-attention to predict the motion features of objects.
This network can help to obtain a more robust node embedding to maintain accurate tracking when objects are on side views of cameras. Then, we further post-process the prediction results with motion propagation and node merging modules.
Finally, the proposed framework will be evaluated with a comprehensive evaluation criterion to demonstrate its robustness compared against previous MC-MOT frameworks. The proposed method even helps to improve the detection accuracy of a standard 3D object detector on the nuScenes benchmark.

\section{Related Work}

MOT problem on AVs has recently received a lot of attention from the research community. There is an increasing amount of research work targeting trajectory estimation on moving sensors \cite{weng2020ab3dmot, chiu2020probabilistic} or combining appearance information to determine object IDs \cite{zhou2019objects, zhou2020tracking, Hu3DT19}.

\paragraph{Tracking using Motion Model}

Weng et al. \cite{weng2020ab3dmot} propose a simple yet effective baseline that utilizes classic state estimator Kalman Filter for 3D bounding boxes. They can be obtained not only from a LiDAR point cloud object detector \cite{Shi_2019_CVPR, 2019arXiv190809492Z, qi2016pointnet, qi2017pointnetplusplus, zhou2017voxelnet} but also from an image-based object detector \cite{Ren17CVPR, zhou2019objects, Simonelli_2019_ICCV, Hu3DT19}. Chiu et al. \cite{chiu2020probabilistic} improves the Kalman Filter tracking system by measuring the Mahalanobis distance between the predicted states and observations. This method is promisingly reliable in filtering outliers and handling both partially and fully occluded objects.

\paragraph{Tracking using Appearance Model}

Zhou et al.'s approaches \cite{zhou2019objects, zhou2020tracking} are widely used in single camera tracking problems. By treating objects as points, these approaches simplify the tracking procedure that is usually a combination of many expensive steps from detection to assigning object ID. Simonelli et al. \cite{Simonelli_2019_ICCV} introduce a novel disentangling transformation for detection loss and a self-supervised term for bounding boxes confidence score. Hu et al. \cite{Hu3DT19} try to estimate robust 3D box information from 2D images then adopt 3D box-reordering and LSTM as a motion module to link objects across frames.

\paragraph{Tracking using Hybrid Approaches}

Chaabane et al. \cite{Chaabane2021deft} train the object detection and the object association task simultaneously by adding a feature extractor and a matching head after object detector. Besides, an LSTM is used as a motion prediction module as an alternative to Kalman Filter. Similarly, Yin et al. \cite{yin2021center} follow the same process, but perform feature extraction on point cloud maps.

\paragraph{Tracking using Modern Approaches}

Graph Neural Network, Self-Attention, and Transformer \cite{vaswani2017attention} introduce a new learning-from-context paradigm. It recently has attracted considerable attention from the research community because of its promising performance in a wide range from Natural Language Processing \cite{ott2018scaling, devlin2019bert, Radford2018ImprovingLU, liu2019roberta} to Computer Vision \cite{dosovitskiy2020, carion2020endtoend, wang2020endtoend, ramachandran2019standalone, touvron2021training, zhu2020deformable} tasks. Currently, there are none of these methods applied in MC-MOT on autonomous vehicles but it is worthy to name a few SCT-MOT approaches \cite{Gao_2019_CVPR, Chu_2017_ICCV, sun2020transtrack, meinhardt2021trackformer, Zhu_2018_ECCV, Weng2020_GNN3DMOT, Weng2020_GNNTrkForecast}. Weng et al. \cite{Weng2020_GNN3DMOT} propose the first feature interaction method that leverages Graph Neural Network to individually adapt an object feature to another object features. Meinhardt et al. \cite{meinhardt2021trackformer} propose a new tracking-by-attention paradigm besides existing tracking-by-regression, tracking-by-detection and tracking-by-segmentation to deal with occlusions and reason out tracker's spatio-temporal correspondences. Sun et al. \cite{Zhu_2018_ECCV} utilize Query-Key mechanism to perform joint-detection-and-tracking, disentangle complex components in previous tracking systems.

\section{Our Proposed Method}

In this section, we first overview our proposed 3D object tracking pipeline where we construct and maintain a Global Graph with the Graph Transformer Networks in Subsection \ref{sec:problem_formulation}.
Then, Subsection \ref{sec:attn_dyn_graph} will detail the structure of Graph Transformer Networks and how it is used to model appearance and motion of tracked objects.
Finally, Subsection \ref{sec:training_gtn} describes how we train the Graph Transformer Networks.

\subsection{MC-MOT via Global Graph Constructing}
\label{sec:problem_formulation}

Given $C$ cameras, denoted by the set $\cC=\{c_1,\dots, c_C\}$, they are used to perceive surrounding environment of a vehicle. In MC-MOT, we assume each camera attached with an off-the-shelf 3D object detector to provide initial location of objects in real-world coordinates. In this work, KM3D \cite{2009.00764} is used to provide 3D object location and features but it can be replaced by any other 3D object detectors.

In the previous MC-MOT approaches \cite{cai2014exploring} \cite{chen2016equalized}, \cite{Zhang2017MultiTargetMT} \cite{Qian_2020_CVPR_Workshops}, the methods depend on tracking results of an MOT algorithm on each camera independently. There is no mechanism to model the relationship between cameras while they have a strong relations.
Instead, our proposed MC-MOT take detection results directly from the detectors and match with current tracked objects using an auto-regressive approach by taking the cameras relation into consideration.

In our approach, a single graph is constructed and maintained across time by graph transformer networks (detailed in Sec. \ref{sec:attn_dyn_graph}).

At time step $t$, our MC-MOT framework receives detection outcomes $ \mathcal{O}_c^{(t)} = \{ \mathbf{o}_{i,c}^{(t)}\}$ generated by a 3D object detector from all synchronized camera inputs. The detected $i$-th object $\mathbf{o}_{i, c}^{(t)}$ contains its location in 3D $\mathbf{l}_{i, c}^{(t)}$ and its features $\mathbf{f}_{i, c}^{(t)}$.
Then, our MC-MOT framework will update and maintain a set of tracked objects, called tracklets $\cT_c^{(t)} = \{\mathbf{tr}^{(t)}_{k,c}\}$, based on detected objects at time step $t$ and previous tracklets at time step $t-1$. Each $\mathbf{tr}^{(t)}_{k,c}$ is a vector with 3D location and features of the corresponding tracked object.
This set of tracklets are represented by a global graph $\cG^{(t)} = (\cV^{(t)}, \cE^{(t)})$, where the vertex set $\cV^{(t)}$ contains \emph{all the tracklets $\cT_c^{(t)}$} tracked up to time $t$ and the edge set $\cE^{(t)}$ contains \textit{geometry distance} between two tracklets.
In this way, $\cG^{(t)}$ can be obtained using graph transformer networks from a joint set of $N_{\mathcal{T}}$ nodes of the previous graph $\cG^{(t-1)}$ and $N_{\mathcal{O}}$ new nodes formed by current detections $ \mathcal{O}_c^{(t)}$s.
The changes in the global graph from frame-to-frame are likely adding new nodes as new objects are detected or removing old nodes as tracklets are out of view. This step is done by graph link prediction using a Softmax classifier similar to \cite{quach2021dyglip}.
Next, we will discuss how the transformer decoder can be employed to update the embedding features for each node with self-attention layer and how to predict tracked objects' motion via cross-attention layer.

\begin{figure*}
    \centering
    \includegraphics[width=0.85\linewidth]{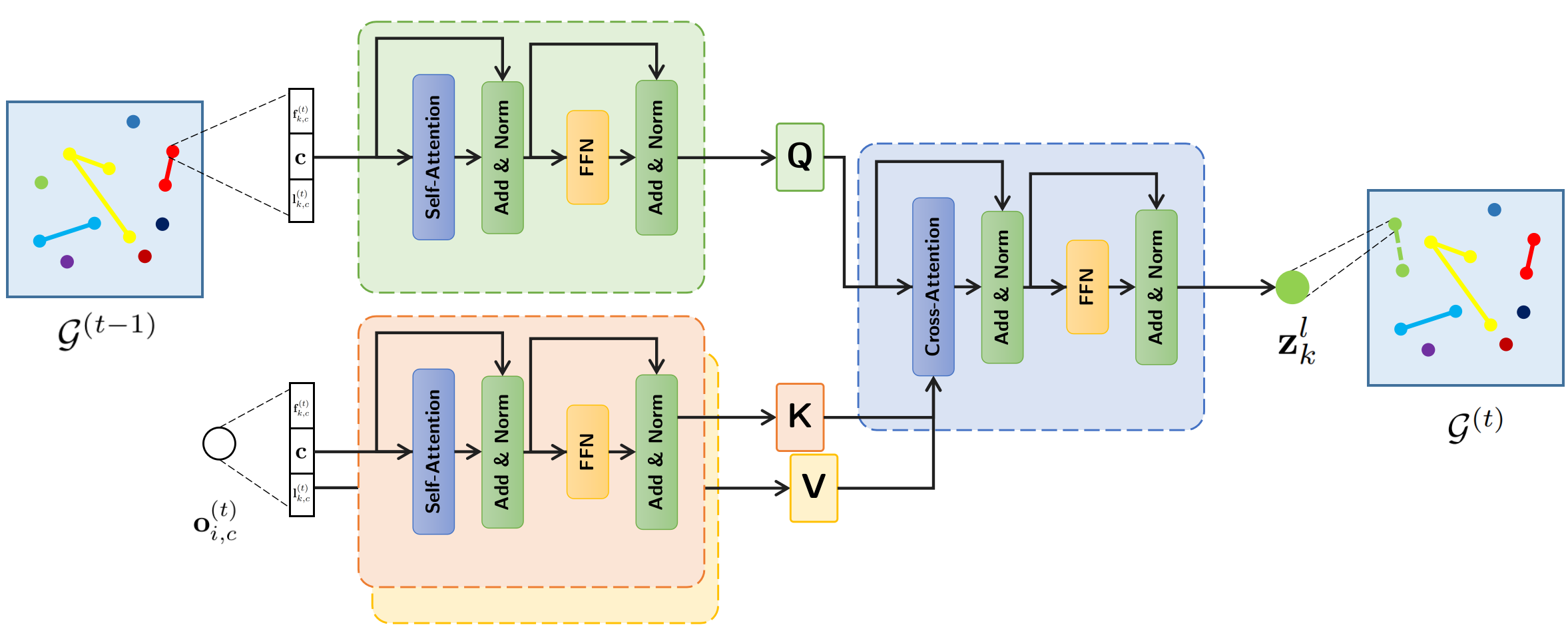}
    \caption{The proposed framework via Graph Transformer Networks. For every new detected object, we calculate new graph feature described in Sub-sec. \ref{ssec:graph_attention} and \ref{ssec:graph_transformer}. Then, we perform motion propagation and node merging operators that include the removing and the adding nodes in the graph via link prediction in Sub-sec. \ref{ssec:graph_propagation} and \ref{ssec:node_merging}.}
    \label{fig:framework}
\end{figure*}

\subsection{Auto-Regressive Graph Transformer Networks}
\label{sec:attn_dyn_graph}

In this section, we introduce Graph Transformer Networks (GTN) to transform and update node embeddings by attending to other nodes for robust appearance and motion modeling.
First, the building blocks of this GTN, i.e. graph self-attention layer and graph cross-attention layer, are presented in Sub-sec. \ref{ssec:graph_attention} and \ref{ssec:graph_transformer}, respectively.
Then, we perform motion propagation and node merging operators that include the removing and the adding nodes in the graph via link prediction in Sub-sec. \ref{ssec:graph_propagation} and \ref{ssec:node_merging}, respectively.

\subsubsection{Graph Self-Attention Layer for Appearance Modeling}
\label{ssec:graph_attention}

Each node $k \in \cV^{(t)}$ in the graph $\cG^{(t)}$ contains the object's 3D location $\mathbf{l}_{k, c}^{(t)}$ and its feature embedding $\mathbf{f}_{k, c}^{(t)}$, i.e. Re-ID features. The Re-ID features are provided by KM3D \cite{2009.00764} as its outputs together with 3D box predictions.
To consider the effects of cameras on appearance features, the self-attention layer takes the input node features as the concatenation of embedding features with camera and location encoding as $ \mathbf{h}^l_k = \{ \mathbf{f}_{k, c}^{(t)} | \mathbf{c} | \mathbf{l}_{k, c}^{(t)} \} \in \mathbb{R}^{D_E}$ , where $ l=0 $ only applied for the input of the first layer, $\mathbf{f}_{k, c}^{(t)} \in \mathbb{R}^{D_F}$, $ \mathbf{c} \in \mathbb{R}^{D_C}$ and $ \mathbf{l}_{k, c}^{(t)} \in \mathbb{R}^{3}$.
We use pre-computed camera and location encoding to concat with the node features before the first layer, similar to how positional encodings are added in the original Transformer \cite{vaswani2017attention}.
Then, the self-attention layer provides the output embeddings as $ \mathbf{h}^{l+1}_k $ for layer $l$. This output can be used as the input for the next layer if there is more than one self-attention layer.

In order to further improve pairwise attention scores as in \cite{vaswani2017attention}, we incorporate pairwise edge features by multiplying them together. In summary, the output of the self-attention layer is computed as follows,
\begin{eqnarray} \label{eqn:str_att_layer}
    \footnotesize
    \mathbf{h'}^{l+1}_{k} = \mathbf{O}_h^l \overset{H}{\underset{i=1}{\Vert}}\left( \sum_{j \in \cV^{(t)}} \mathbf{w}^{i,l}_{kj} \mathbf{V}^{i,l} \mathbf{h}^l_k  \right) \\
    \mathbf{e'}^{l+1}_{kj} = \mathbf{O}_e^l \overset{H}{\underset{i=1}{\Vert}}\left( \sum_{j \in \cV^{(t)}} \mathbf{w'}^{i,l}_{kj} \right) \\
    \mathbf{w}^{i,l}_{kj} = \text{softmax}_j ( \mathbf{w'}^{i,l}_{kj} ) \\
    \mathbf{w'}^{i,l}_{kj} = \left( \frac{\mathbf{Q}^{i,l} \mathbf{h}^l_k \cdot \mathbf{K}^{i,l} \mathbf{h}^l_j }{\sqrt{D_h}} \right) \cdot \mathbf{E}^{i,l} \mathbf{e}^{l}_{kj}
\end{eqnarray}
where $\mathbf{w}^{i,l}_{kj}$ are the attention coefficients for the $i$-th attention head, $\Vert$ is the feature vector concatenation operation,
$\mathbf{Q}^{i,l}, \mathbf{K}^{i,l}, \mathbf{V}^{i,l}, \mathbf{E}^{i,l} \in \mathbb{R}^{D_Z \times D_{E}}$ denote the ``queries", ``keys", ``values" linear projection matrices and node embedding, respectively, as defined in \cite{vaswani2017attention} and $D_Z$ is the output feature dimension. $H$ denotes number of attention head in multi-head attention setting.  

The outputs $ \mathbf{h}^{l+1}_{k} $ and $\mathbf{e}^{l+1}_{kj}$ are then passed through feed forward layers with residual connections and normalization layers (see Fig. \ref{fig:framework}), defined as follows.
\begin{eqnarray}
    \mathbf{h''}^{l+1}_{k} = \text{norm} \left( \mathbf{h'}^{l+1}_{k} + \mathbf{h}^{l}_{k}   \right) \\
    \mathbf{h'''}^{l+1}_{k} = \text{FFN}^l_h \left( \mathbf{h''}^{l+1}_{k} \right) \\
    \mathbf{h}^{l+1}_{k} = \text{norm} \left( \mathbf{h''}^{l+1}_{k} + \mathbf{h'''}^{l+1}_{k}   \right)
\end{eqnarray}
where $\mathbf{h''}^{l+1}_{k}$ and $\mathbf{h'''}^{l+1}_{k}$ denote the outputs of intermediate layers. FFN is the feed forward layers.
\begin{eqnarray}
    \mathbf{e''}^{l+1}_{kj} = \text{norm} \left( \mathbf{e'}^{l+1}_{kj} + \mathbf{e}^{l}_{kj}   \right) \\
    \mathbf{e'''}^{l+1}_{kj} = \text{FFN}^l_e \left( \mathbf{e''}^{l+1}_{kj} \right) \\
    \mathbf{e}^{l+1}_{kj} = \text{norm} \left( \mathbf{e''}^{l+1}_{kj} + \mathbf{e'''}^{l+1}_{kj}   \right)
\end{eqnarray}
where $\mathbf{e''}^{l+1}_{k}$ and $\mathbf{e'''}^{l+1}_{k}$ denote the outputs of intermediate layers.

\subsubsection{Graph Transformer Layer for Motion Modeling}
\label{ssec:graph_transformer}

In this section, we demonstrate how tracked objects in tracklet nodes are used as queries while newly detected objects are used as keys and values in our proposed transformer layer.
This layer perform a cross-attention mechanism instead of self-attention mechanism where queries are different from keys.
The input of this layer are the output node embedding from previous self-attention layers and the output of this layer are new tracklet nodes for the current frame $t$.
It takes an object feature from previous frames as input query instead. This inherited object feature conveys the appearance and location information of previously seen objects, so this layer could well locate the position of the corresponding object on the current frame and output “tracking boxes”.
This design helps to capture the attention on current frame detection features and previous frame track queries, to continuously update the representation of object identity and location in each track query embedding. 

We first put together all detected objects as $X_{\mathcal{O}} \in \mathbb{R}^{N_{\mathcal{O}} \times D_Z}$ and all tracked objects as $X_{\mathcal{T}} \in \mathbb{R}^{N_{\mathcal{T}} \times D_Z}$.
Then the $l$-th output of the multi-head cross attention layer is defined as
\begin{eqnarray} \label{eqn:cross_att_layer}
    \small
    \mathbf{z}^{l}_{k} = \mathbf{O}_z^l \overset{H}{\underset{i=1}{\Vert}}\left( \sum_{j \in \mathbf{X}_{\mathcal{O}}} \mathbf{W}^{i,l}_{kj} \mathbf{V}^{i,l} \mathbf{X}^T_{\mathcal{T}}[k]  \right) \\
    \mathbf{W}^{i,l}_{kj} = \text{softmax}_j \left( \frac{\mathbf{Q}^{i,l} \mathbf{X}^T_{\mathcal{T}}[k]  \cdot \mathbf{K}^{i,l} \mathbf{X}^T_{\mathcal{O}}[j] }{\sqrt{D_h}}  \right)
\end{eqnarray}
where $\mathbf{Q}^{i,l}, \mathbf{K}^{i,l}, \mathbf{V}^{i,l} \in \mathbb{R}^{D_E \times D_{Z}}$, are the ``queries", ``keys" and ``values" linear projection matrices, respectively, as defined in \cite{vaswani2017attention} and $D_Z$ is the output feature dimension.

Similar to attention layer, we can stack multiple cross-attention layers together. Then we get the final output to pass through FFN to provide final set of new node embeddings including location and class predictions for frame $t$.

\subsubsection{Cross-Camera Motion Propagation}\label{ssec:graph_propagation}

In this section, we provide a more detailed formulation on how to obtain Re-ID features of the detected objects from camera $c_k$ to camera $c_j$.
First, we compute the transformation matrix to transform 3D object locations to 2D/image coordinates. This transformation which is composed of a transformation from camera-to-world for camera $c_k$, a transformation from world-to-camera for camera $c_j$, and a transformation from camera-to-image for camera $c_j$, is defined as follows.
\begin{equation}
    \mathbf{M}_{kj} = \mathbf{M}_{I_j} * \mathbf{M}_{E_j} * \mathbf{M}^{-1}_{E_k}
\end{equation}
where $\mathbf{M}_{E_j}$ and $\mathbf{M}^{-1}_{E_k}$ are the extrinsic camera matrix for camera $c_k$ to camera $c_j$, respectively. $\mathbf{M}_{I_j}$ is the intrinsic camera matrix for camera $c_j$.
Note that we only consider two adjacent cameras $c_k$ and $c_j$ where they have a certain amount of overlapping views.
Then, we use the transformed 2D/image location to extract the re-id features at the corresponding location on the image.
Finally, we update the existing node or add a new node for all the tracked objects $\mathbf{tr}^{(t)}_{k,c_j}$.

\subsubsection{Node Merging via Edge Scoring}
\label{ssec:node_merging}
After having transformed node and edge features, we train a fully connected
layer and a softmax layer as a classifier to determine the similarity between two nodes as previously proposed in \cite{quach2021dyglip}. The classifier produces a probability score $s \in [0, 1]$. The higher the score is, the more likely the two nodes are linked. Then we remove detection nodes that have a low class score which indicates that the detection is matched with an existing tracklet.
We also merge nodes that have high similarity scores that have the same camera encoding, i.e. detected within single camera and update edge weights as the similarities among tracklet nodes to indicate the same target ID from different cameras.
These necessary steps are similar to a non-maximum suppression (NMS) applied to trajectory for post-processing although cross-attention layer help spatially discriminate almost identical track query embeddings merging to the same target ID.

\subsection{Processing Flow}

In this section, we briefly summarize the pipeline of our proposed graph transformer networks to predict tracklet motion, motion propagation and node merging in Algorithm \ref{alg:DyGLIP}.

\begin{algorithm}[th]
    \caption{The process pipeline for global graph constructing, motion prediction, propagation \& node merging}
    \label{alg:DyGLIP}
    \begin{algorithmic}[1]
        \STATE Init $t\gets 0$ /* Time */, $V \gets \emptyset$
        \WHILE{$t < t_{\max}$}
        \STATE Obtain the set of detected objects $\mathcal{O}_c^{(t)}$ from 3D object detector \cite{2009.00764} in all cameras.
        \FOR{$\mathbf{o}_{k, c}^{(t)} \in \mathcal{O}_c^{(t)}$}
        \STATE $\cV^{(t)} \gets \cV^{(t-1)} \cup \mathbf{o}_{k, c}^{(t)}$ /* Add new nodes to graph */
        \STATE /* Use the vector $\{ \mathbf{f}_{k, c}^{(t)} | \mathbf{c} | \mathbf{l}_{k, c}^{(t)} \}$ as node features. */
        \ENDFOR
        \FOR{$k \in \cV^{(t)}$}
        \STATE Obtain new node embedding $\mathbf{h'}_{k}$ /* Section 3.2.1 */
        \ENDFOR
        \STATE Obtain new set of nodes $\cV'^{(t)}$ with location and classification of tracked objects $\mathbf{tr}^{(t)}_{k,c}$ via motion modeling /* Section 3.2.2 */
        \FOR{$c \in C$}
        \STATE Propagate the location of $\mathbf{tr}^{(t)}_{c}$ to adjacent cameras /* Section 3.2.3 */
        \ENDFOR

        \FOR{$v_i \in \cV'^{(t)}$}
        \STATE Obtain edge scoring  to the remaining nodes and node merging /* Section 3.2.4 */
        \STATE Assign ID based on edge scores.
        \ENDFOR
        \STATE $t \gets t + 1$
        \ENDWHILE

    \end{algorithmic}
\end{algorithm}

\subsection{Model Training}
\label{sec:training_gtn}

In this section, we present how to train our proposed graph transformer networks, including self-attention and cross-attention layers.

\paragraph{Training Data.}

We train our proposed method on a large-scale dataset, i.e. nuScenes, training set with 750 scenes of 20s each and use its validation set for our ablation study.
The ground truth 3D bounding boxes and the extracted ReID features from the pre-trained models in \cite{zhou2019osnet, Qian_2020_CVPR_Workshops} were used together as the inputs for training GTN.
Each training sample contains a chunk size of two consecutive frames from a training sequence.

\paragraph{Training Loss.}

Our framework can be trained with two adjacent frames by optimizing for detections and tracklets prediction at frame $t$, given previous frame tracklets.
Our joint objective function include \textit{learning node embedding} capturing both structural information from the graph, \textit{computing weighted linking score} between two nodes in the graph and \textit{learning to predict tracklets motion}.

For \textit{learning node embedding}, we measure binary cross-entropy loss $ \mathcal{L}_{emb} $ between nodes that belong to the same objects for the model to output similar feature embeddings.
\begin{equation}
    \footnotesize
    \begin{split}
        \mathcal{L}_{emb}(v_k) = & \sum_{v_j \in \cN_b^{(t)}(v_k)}  -\log
        \left( \sigma \left( < e'_{v_k} , e'_{v_j} > \right) \right)  \\
        & - w_g \sum_{v_i \in \cN_g^{(t)}(v_k)} \log \left( 1 - \sigma \left( < e'_{v_k} , e'_{v_i} > \right) \right) \\
    \end{split}
\end{equation}
where $ <\cdot> $ is the inner production between two vectors, $\sigma$ is Sigmoid activation function, $\cN_b^{(t)}(v_k)$ is the set of fixed-length random walk neighbor nodes of $v_k$ at time step $t$, $\cN_g^{(t)}(v_k)$ is a negative samples of $v_i$ for time step $t$,  $\cN_a^{(t)}(v_k) = \cN_b^{(t)}(v_k) \cup \cN_g^{(t)}(v_k)$ and $w_g$, negative sampling ratio, is an adjustable hyper-parameter to balance the positive and negative samples.

For edge scoring, we use a cross-entropy loss function $ \mathcal{L}_c (e_{kj}) $ based on measurement features to ensure the score between two nodes that are connected is higher than other nodes.

For \textit{learning to predict tracklets motion}, we set prediction loss to measure the set of predictions for $N_{\mathcal{O}}$ detections and $N_{\mathcal{T}}$ tracklets comparing with  ground truth objects in terms of classification and location (bounding boxes).
Set-based loss produces an optimal bipartite matching between $N_{\mathcal{O}}$ detections and ground truth objects while $N_{\mathcal{T}}$ tracklets will be matched with boxes from previous frames. The matching cost is defined as follows.
\begin{equation}
    \mathcal{L}_{set} = \overset{N_{\mathcal{O}} + N_{\mathcal{T}}}{\underset{i=1}{\sum}} \left( \lambda_{cls} \mathcal{L}_{cls} + \lambda_{box} \mathcal{L}_{box} + \lambda_{iou} \mathcal{L}_{iou} \right)
\end{equation}
where $ \lambda_{cls}, \lambda_{box}$ and $\lambda_{iou}$ are combination weighting parameters for each component losses. $\mathcal{L}_{cls}$ is the cross-entropy loss between prediction classification and ground truth category labels. $\mathcal{L}_{box}$ and $\mathcal{L}_{iou}$ are the $\ell_1$ loss and the generalized intersection over union (IoU) \cite{rezatofighi2019generalized} for 3D bounding boxes.
Finally, we have the total loss defined as
\begin{equation}
    \mathcal{L}_{total} = \mathcal{L}_{emb} + \mathcal{L}_{c} + \mathcal{L}_{set}
\end{equation}

\begin{table}[!t]
    \small
    \centering
    \begin{tabular}{|c|c|c|c|c|}
        \hline
        \textbf{Method}              & \textbf{mATE} $\downarrow$ & \textbf{mASE} $\downarrow$ & \textbf{mAOE} $\downarrow$ & \textbf{mAVE} $\downarrow$ \\
        \hline
        3D KF \cite{weng2020ab3dmot} & 0.8153                     & 0.5155                     & 0.7382                     & 1.6186                     \\
        LSTM \cite{Chaabane2021deft} & 0.8041                     & 0.4548                     & 0.6744                     & 1.6139                     \\
        \hline
        \textbf{Ours}                & \textbf{0.5132}            & \textbf{0.4388}            & \textbf{0.3677}            & \textbf{1.2189}            \\
        \hline
    \end{tabular}
    \caption{Motion Errors comparison for different motion modeling}
    \label{tab:motion_errors}
\end{table}

\section{Experimental Results}

In this Section, we detail the benchmark dataset and metrics in Subsection \ref{ssec:data_metrics}. Then, the setups for all experiments and the ablation study will be presented in Subsections \ref{ssec:exp_setup} and \ref{ssec:ablation_study} respectively. The comparisons with the State-of-the-Art (SOTA) methods will be detailed in Subsection \ref{ssec:compare_results} on a large-scale Tracking Challenge, i.e. nuScenes Vision Track.

\subsection{Benchmark Dataset and Metrics}
\label{ssec:data_metrics}

\subsubsection{Dataset}
\paragraph{nuScenes} \cite{caesar2020nuscenes} is one of the large-scale datasets for Autonomous Driving with 3D object annotations. It contains 1,000 videos of 20-second shots in a setup of 6 cameras, i.e. 3 front and 3 rear ones,  with a total of 1.4M images. It also provides 1.4M manually annotated 3D bounding boxes of 23 object classes based on LiDAR data. This dataset is an official split of 700, 150 and 150 videos for training, validation and testing, respectively.

\subsubsection{Metrics}

The proposed method is evaluated using both detection and tracking metrics described in \cite{caesar2020nuscenes}.

\paragraph{Detection Metrics.} A commonly used metric, i.e. \textit{Mean Average Precision (mAP)}, is defined as a match using a 2D center distance on the ground plane instead of intersection over union cost for nuScenes detection challenges.

Similarly, other motion-related metrics are also defined in nuScenes, such as \textit{Average Translation Error (ATE)} measuring Euclidean center distance in 2D in meters, \textit{Average Scale Error (ASE)} computing as $1 - IOU$ after aligning centers and orientation, \textit{Average Orientation Error (AOE)} measuring by the smallest yaw angle difference between prediction and ground-truth in radians, \textit{Average Velocity Error (AVE)} measuring the absolute velocity error in $m/s$ and \textit{Average Attribute Error (AAE)} computing as $1 - acc$, where $acc$ is the attribute classification accuracy.

\begin{figure*}[!t]
    \centering
    \includegraphics[width=0.8\linewidth]{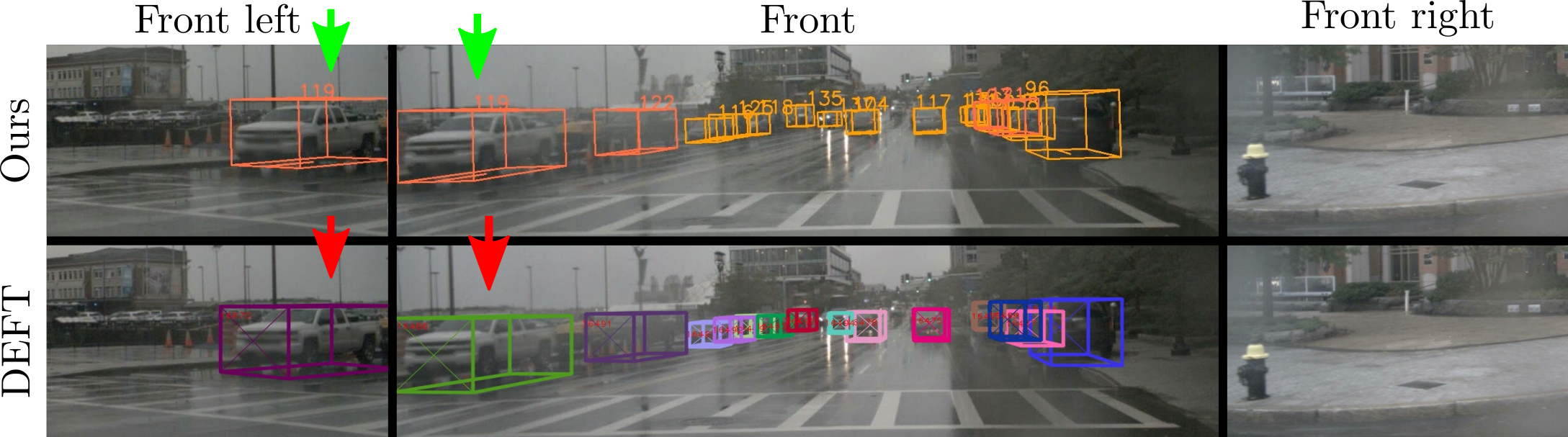}
    \caption{Our proposed method (top) can recognize a positive tracking case compare with a MC-MOT system which has no object's correlations linking module (i.e. DEFT) for all cameras (bottom). Green arrows indicate true positive tracking samples, red arrows indicate false negative tracking samples. Best viewed in color and zoom in.}
    \label{fig:compare_tracking}
\end{figure*}

Last but not least, we also use the \textit{nuScenes Detection Score (NDS)} that is based on a simple additive weighting of the mean of all other metrics above, including \textit{mAP}, \textit{mATE}, \textit{mASE}, \textit{mAOE}, \textit{mAVE} and \textit{mAAE}.

\paragraph{Tracking Metrics.}  The tracking performance is measured using the popular \textit{CLEAR MOT} metrics \cite{bernardin2008evaluating} including \textit{MOTA}, \textit{MOTP}, ID switch (\textit{IDS}), mostly tracked (\textit{MT}), mostly lost (\textit{ML}), fragmented (\textit{FRAG}). Similar to nuScenes, we use two accumulated metrics introduced in \cite{weng2020ab3dmot} as the main metrics, including the average over the MOTA metric (\textit{Average MOTA (AMOTA)}) and the average over the MOTP metric (\textit{Average MOTP (AMOTP)}).

\subsection{Experiments Setup}
\label{ssec:exp_setup}

The proposed graph transformer networks module is trained with two consecutive frames where the graph $ \{ \cG^{(t - 1)} \} $ in the previous time step is used to predict new graph $ \cG^{(t)} $ at time step $t$.
Then, Mini-batch (chunk of two) gradient descent is employed with Adam optimizer to learn all the parameters in the attention layers.

\subsection{Ablation Study}
\label{ssec:ablation_study}

In this section, we present some experiments to ablate the effect of each component of the proposed framework.  Particularly, this section aims to demonstrate the followings: 1. better motion modeling with cross-attention layer in GTN; 2. the role of architecture choice of graph transformer networks.

\begin{table}[!t]
    \footnotesize
    \centering
    \begin{tabular}{|c|c|c|c|c|c|c|c|}
        \hline
        \textbf{Structures}         & \textbf{mATE} $\downarrow$ & \textbf{mASE} $\downarrow$ & \textbf{mAOE} $\downarrow$ & \textbf{mAVE} $\downarrow$ \\
        \hline
        Self-attn 1-layer           & 0.812                      & 0.298                      & 0.820                      & \textbf{1.187}             \\
        Self-attn 2-layer           & 0.785                      & \textbf{0.286}             & 0.703                      & 1.284                      \\
        \textbf{Self-attn 3-layer}  & \textbf{0.750}             & 0.293                      & \textbf{0.485}             & 1.432                      \\
        \hline
        Cross-attn 1-layer          & 0.824                      & 0.293                      & 0.866                      & 1.281                      \\
        Cross-attn 2-layer          & 0.772                      & \textbf{0.279}             & 0.670                      & 1.287                      \\
        \textbf{Cross-attn 3-layer} & \textbf{0.513}             & 0.439                      & \textbf{0.368}             & \textbf{1.219}             \\
        \hline
    \end{tabular}
    \caption{Ablation study on different configuration for self-attention and cross-attention layers.}
    \label{tab:attn_structures}
\end{table}

\begin{figure*}[!t]
    \centering
    \includegraphics[width=0.8\linewidth]{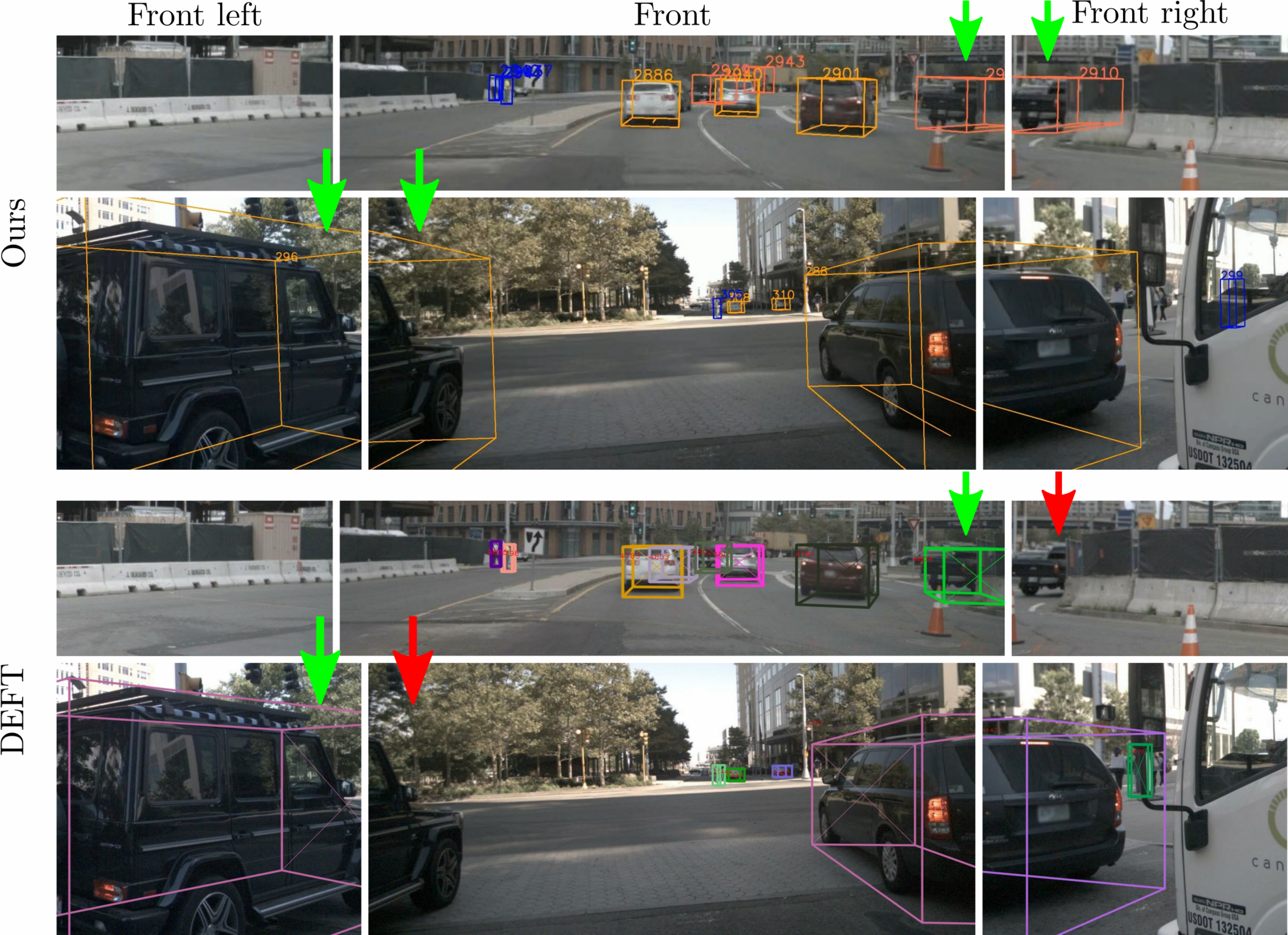}
    \caption{Our proposed method (top) can recover a false negative detection case compared with a MC-MOT system which runs independently on each camera (i.e. DEFT) (bottom). Green arrows indicate true positive detection samples, red arrows indicate false negative detection samples. Best viewed in color and zoom in.}
    \label{fig:compare_detection}
\end{figure*}

\begin{table*}[!t]
    \small
    \centering
    \resizebox{1.0\textwidth}{!}{
        \begin{tabular}{|l|c|c|c|c|c|c|c|c|c|c|c|c|}
            \hline
            \textbf{Method}                     & \textbf{Glo. Assoc.} & \textbf{AMOTA} & \textbf{AMOTP} & \textbf{MOTAR} & \textbf{MOTA} $\uparrow$ & \textbf{MOTP} $\downarrow$ & \textbf{RECALL} $\uparrow$ & \textbf{MT} $\uparrow$ & \textbf{ML} $\downarrow$ & \textbf{IDS} $\downarrow$ & \textbf{FRAG} $\downarrow$ \\	
            \hline
            MonoDIS \cite{Simonelli_2019_ICCV}  & \xmark               & 0.045          & 1.793          & 0.202          & 0.047                    & 0.927                      & 0.293                      & 395                    & 3961                     & 6872                      & 3229                       \\ 
            CenterTrack \cite{zhou2020tracking} & \xmark               & 0.068          & 1.543          & 0.349          & 0.061                    & 0.778                      & 0.222                      & 524                    & 4378                     & 2673                      & 1882                       \\ 
            DEFT \cite{Chaabane2021deft}        & \xmark               & 0.213          & 1.532          & 0.49           & 0.183                    & 0.805                      & 0.4                        & 1591                   & 2552                     & 5560                      & 2721                       \\ 
            QD-3DT \cite{Hu2021QD3DT}           & \xmark               & \textbf{0.242} & 1.518          & \textbf{0.58}  & \textbf{0.218}           & 0.81                       & 0.399                      & 1600                   & 2307                     & 5646                      & 2592                       \\
            \hline

            \textbf{Ours}                       & \cmark               & 0.24           & \textbf{1.52}  & 0.568          & 0.197                    & \textbf{0.832}             & \textbf{0.453}             & \textbf{1643}          & \textbf{2162}            & \textbf{1362}             & \textbf{1462}              \\

            \hline
        \end{tabular}
    }
    \caption{Comparison of 3D tracking performance on the nuScenes validation set for Vision Track challenge. \textbf{Glo. Assoc.} indicates method linking object IDs across all cameras}
    \label{tab:nuscene_track_results}
\end{table*}

\begin{table*}[!t]
    \centering
    \resizebox{0.75\textwidth}{!}{
        \begin{tabular}{|l|c|c|c|c|c|c|c|c|}
            \hline
            \textbf{Method}                                            & \textbf{mAP} $\uparrow$ & \textbf{NDS} $\uparrow$ & \textbf{mATE} $\downarrow$ & \textbf{mASE} $\downarrow$ & \textbf{mAOE} $\downarrow$ & \textbf{mAVE} $\downarrow$ & \textbf{mAAE} $\downarrow$ \\
            \hline
            MonoDIS \cite{Simonelli_2019_ICCV}                         & 0.2976                  & 0.3685                  & 0.7661                     & 0.2695                     & \textbf{0.5839}            & 1.3619                     & 0.184                      \\
            MonoDIS \cite{Simonelli_2019_ICCV} +  \textbf{Our MP + NM} & \textbf{0.3019}         & \textbf{0.3893}         & \textbf{0.6558}            & \textbf{0.2410}            & 0.6787                     & \textbf{1.3209}            & \textbf{0.184}             \\
            \hline
            \hline
            CenterNet \cite{zhou2019objects}                           & 0.3027                  & 0.3262                  & 0.7152                     & 0.2635                     & \textbf{0.6158}            & 1.4254                     & 0.6567                     \\
            CenterNet \cite{zhou2019objects} +  \textbf{Our MP + NM}   & \textbf{0.3487}         & \textbf{0.4016}         & \textbf{0.5417}            & \textbf{0.2023}            & 0.6317                     & \textbf{1.3094}            & \textbf{0.6567}            \\
            \hline
            \hline
            KM3D \cite{2009.00764}                                     & 0.2763                  & 0.3201                  & 0.7495                     & 0.2927                     & 0.4851                     & \textbf{1.4322}            & 0.6535                     \\
            KM3D \cite{2009.00764} + \textbf{Our MP + NM}              & \textbf{0.3503}         & \textbf{0.4117}         & \textbf{0.6998}            & \textbf{0.2323}            & \textbf{0.1861}            & 1.8341                     & \textbf{0.5166}            \\
            \hline
        \end{tabular}
    }
    \caption{Comparison of 3D object detectors with and without using our motion propagation (MP) and node merging (NM) modules in terms of detection metrics on the nuScenes validation set for Vision Detection challenge}
    \label{tab:nuscene_detection_results}
\end{table*}

\paragraph{The Role of Motion Model}

In this experiment, we evaluate the effectiveness of different motion modeling methods on detection performance. We use the locations predicted by motion models to compare with ground truth locations in terms of motion-related metrics. In such way, we can evaluate how good the motion model capturing and predicting the motion of tracked objects.  We compare with two other commonly used motion models, i.e. 3D Kalman Filter \cite{weng2020ab3dmot} and LSTM \cite{Chaabane2021deft}. As shown in Table \ref{tab:motion_errors}, our GTN gives better results than a classical object state prediction technique, i.e. 3D Kalman Filter used in \cite{weng2020ab3dmot} and a deep learning based technique, i.e. LSTM module, used in \cite{Chaabane2021deft}.

\paragraph{The Configuration for Graph Transformer Networks}

We conduct additional ablation studies to evaluate the effects on configuration of the attention modules in GTN, including the number of attention layers. Table \ref{tab:attn_structures} shows the performance of our proposed framework in terms of detection metrics using various configuration of the attention modules.
We change the number of layer for self-attention and the cross-attention layers independently.
We use a fixed number of layers, i.e. 2, for self-attention and the cross-attention layers while changing the other, respectively.

\subsection{Comparison against The State-of-the-Art Methods}
\label{ssec:compare_results}

In this section, we first compare our proposed framework with other vision-based (without using LiDAR or RADAR information) tracking approaches, which are the top in nuScenes vision only tracking challenge leaderboard. Then we conduct an experiment to demonstrate that using tracked 3D bounding boxes from our tracking framework can actually improve the detection metrics.

\paragraph{Comparison against Tracking Methods on Tracking Metrics}

This experiment compares our proposed method with other vision-based methods, including MonoDIS \cite{Simonelli_2019_ICCV}, CenterTrack \cite{zhou2020tracking} and DEFT \cite{Chaabane2021deft}, QD-3DT \cite{Hu2021QD3DT} which are the top/winner of nuScenes vision only tracking challenge. As we can see in Table \ref{tab:nuscene_track_results}, our method decreases error rates compared to top approaches, i.e. DEFT, in most of the metrics. Fig. \ref{fig:compare_tracking} illustrates the key factor that help improve the tracking performance is that we perform appearance matching across cameras in addition to motion modeling. It shows that our proposed method (top) can assign object ID globally between cameras compared with DEFT \cite{Chaabane2021deft} (bottom).

\paragraph{Comparison against Detection Methods on Detection Metrics}

Table \ref{tab:nuscene_detection_results} demonstrates that the combination of object detector and our motion propagation (MP) and node merging (NM) modules achieves the better results than original object detector. In this experiment, we compare three different 3D object detectors, including KM3D \cite{2009.00764},  MonoDIS \cite{Simonelli_2019_ICCV} and CenterNet \cite{zhou2019objects}.
The best result achieves with the combination of KM3D object detector \cite{2009.00764} and our MP+NM modules since it is guided by global decoded locations from our transformation procedure as described in \ref{ssec:graph_propagation}. Fig. \ref{fig:compare_detection} illustrates the improvement on detector fail cases with the help from our tracking framework.

\section{Conclusions}

This paper has introduced a new global association graph model to solve the MC-MOT problem for AV. The proposed framework can learn to perform tracking frame-by-frame in an end-to-end manner starting from detections to motion prediction and global association tracklets with detections.
These tasks are enhanced with self-attention and cross-attention layers so that the proposed graph can capture both structural and motion across cameras.
The experiments show performance improvements in a large-scale dataset in AV in terms of vision-based detection and tracking accuracy.

\section*{Acknowledgment}

This material is based upon work supported in part by the US NSF Data Science, Data Analytics that are Robust and Trusted (DART) and NSF WVAR-CRESH Grant.

    {\small
        \bibliographystyle{ieee_fullname}
        \bibliography{ref}
    }

\end{document}